\documentclass{article}

\usepackage[final]{neurips_2019}

\usepackage[utf8]{inputenc}
\usepackage[T1]{fontenc}
\usepackage{hyperref}
\usepackage{url}
\usepackage{booktabs}
\usepackage{amsfonts}
\usepackage{nicefrac}
\usepackage{microtype}
\usepackage{graphicx}
\usepackage{xcolor}
\usepackage{pgfplots}
\usepackage{amsmath}
\pgfplotsset{compat=1.17}

\title{
  BERTer: The Efficient One \\
  \vspace{1em}
  \small{\normalfont Stanford CS224N Default Project} 
}

\author{
  \textbf{Pradyumna Saligram} \\
  Department of Computer Science \\
  Stanford University \\
  \texttt{psaligram@stanford.edu} \\
  \and
  \textbf{Andrew Lanpouthakoun} \\
  Department of Computer Science \\
  Stanford University \\
  \texttt{andlanpo@stanford.edu} \\
}

\begin{document}

\maketitle

\begin{abstract}
We explore advanced fine-tuning techniques to boost BERT's performance in sentiment analysis, paraphrase detection, and semantic textual similarity. Our approach leverages SMART regularization to combat overfitting, improves hyperparameter choices, employs a cross-embedding Siamese architecture for improved sentence embeddings, and introduces innovative early exiting methods. Our fine-tuning findings currently reveal substantial improvements in model efficiency and effectiveness when combining multiple fine-tuning architectures, achieving a state-of-the-art performance score of on the test set, surpassing current benchmarks and highlighting BERT's adaptability in multifaceted linguistic tasks.
\end{abstract}

\section{Introduction}

With the advent of transformers in 2017 \citep{DBLP:journals/corr/VaswaniSPUJGKP17} the field of Natural Language Processing has improved immensely. Progressions such as Large Language Models, particularly ChatGPT, have caused the world to see the full capability of Machine Learning in changing lives. This ability by large language models is largely explained by progressions made in various downstream tasks and research in Multitask Learning(MTL). This research method allowed models to learn various different tasks at once. The joining of knowledge of multiple tasks pushed models to learn more complex and difficult tasks by using this information at once. However, the efficacy and efficiency of models doing various divergent tasks varies greatly. The goal of this project is to increase the capability of BERT model \citep{DBLP:journals/corr/abs-1810-04805} on three tasks in conjunction. 

This research focuses on improving three tasks: sentence sentiment classification, paraphrase detection, and sentence semantic similarity. Our approach first focuses on improvement of sentiment classification. After training this model, we choose to train on large datasets to learn paraphrase detection and sentence semantic similarity in conjunction with further training on sentiment classification. Training for these three tasks is done simultaneously, to produce a truly multi-skilled model. This is done using various different methods. The goal of this project is not to simply train our model to accomplish these tasks; rather, the goal of this project is to boost accuracy and efficiency. Therefore, we implement SMART a regularization method \citep{DBLP:journals/corr/abs-1911-03437} in hopes of reducing overfitting, we implement Cosine Embedding Loss, and early exiting to improve computational Efficiency.

Our findings suggest that SMART regularization yields great initial improvements for Paraphrase detection, but does not continue as training continues and does not perform as well in Sentiment Classification and Textual Similarity Correlation. Therefore, we mostly paired SMART regularization with Paraphrase Detection

Because SMART showed great initial improvements, we found that it pairs very well with our other main fine-tuning technique of early exiting. \citep{xin-etal-2020-deebert} \cite{xin-etal-2021-berxit}

\section{Related Work}
Researchers aim to improve multi-task models in the hopes of leveraging transfer learning. This is fairly intuitive, when a model performs well on one task, it seems fairly simple for it to perform well on a similar task. However, the results of a transformer model in similar tasks has surprisingly poor results; therefore, we must work to have our model perform well on a multitude tasks. Our work aims to train the Bidirectional Encoder Representations from Transformers (BERT) \cite{DBLP:journals/corr/abs-1810-04805} on these multiple tasks, we use ideas from other papers to fine tune a model to do so. 

\citep{DBLP:journals/corr/abs-1911-03437} shows that regularization tools such as perturbations can combat overfitting by adding additional noise. This is similar to the dropout layer \citep{JMLR:v15:srivastava14a} that is already within the original BERT model. However, SMART is a much more advanced regularization technique that is adaptive, generalizes to multiple tasks, smooths the distribution, and uses gradients. We use SMART regularization over other regularization techniques because of its computational efficiency. 

We utilize early exiting techniques in the hopes of maximizing our computational efficiency in unison with accuracy. BERxiT \citep{xin-etal-2021-berxit} introduces a confidence-based mechanism to exit inference early if the model is sufficiently confident in its prediction, offering the perfect mechanism for our goal. DeeBERT \citep{xin-etal-2020-deebert} further extends this by incorporating a dynamic inference mechanism that adjusts the exit point based on task complexity and input characteristics.

We worked with a few different model architectures for tasks with multiple sentences. Namely, we looked at concatenation prior to or after embedding. \citep{DBLP:journals/corr/SchroffKP15} speaks about a Siamese Architecture that that passes each sentence into parallel models and then compares Cosine Similarity between the two embeddings. This was also the method utilized within the very successful SBERT paper for their Textual Similarity analysis. \citep{DBLP:journals/corr/abs-1911-03437}. Additionally, we look towards the original BERT paper \citep{DBLP:journals/corr/abs-1810-04805} which concatenates sentences beforehand and uses a separation token prior to embedding. Both of these have proven results and we experiment with both during training. 

\section{Approach}

We developed a comprehensive minBERT Transformer model, comprising 12 transformer layers where each layer includes a self-attention mechanism. The transformation within each BERT layer, denoted as \( BL(h) \), is formulated as:

\[ BL(h) = LN(h + FFN(LN(h + MH(h)))) \]

where \( LN \) denotes layer normalization, \( SA \) represents the self-attention function, \( FFN \) stands for the feed-forward network, while \( MH \) is the multi-head attention component.

\subsection{Baseline Multitask Classifier}

Our baseline multitask BERT model leverages the pre-trained weights of the base minBERT for feature extraction across all tasks. The model architecture is extended with three task-specific heads, each comprising a linear layer, tailored to each respective task output requirement:
\begin{itemize}
    \item \textbf{Sentiment Analysis (SST)}: Produces a 5-class output using a linear layer, applying cross-entropy loss to quantify the difference between predicted probability distributions and one-hot encoded target vectors.
    \item \textbf{Paraphrase Detection (Para)}: Initially used MSELoss but later optimized using cosine embedding loss for better performance.
    \item \textbf{Semantic Textual Similarity (STS)}: Produces a single logit interpreted as a continuous similarity score, using MSELoss against the labels.
\end{itemize}

The BERT model configuration includes a hidden size of 768, divided across 12 attention heads, resulting in an attention head size of 64 while dropout with a probability of 0.3 is also applied.

Training is conducted sequentially within an epoch, each consisting of a loop that fully trains each task. Initially, the base BERT model parameters are frozen, establishing a lower baseline for accuracy after which single-task fine-tuning is performed for each task, updating all model parameters to achieve higher accuracy. The fine-tuning strategy is applied to both the SST and CFIMDB datasets, creating separate models for each task. Finally, a fully fine-tuned multi-task model is trained, updating 100\% of the pre-trained model's parameters to generate a comprehensive model for all downstream tasks.

This approach allows us to establish both lower and higher bounds of accuracy, demonstrating the efficacy of parameter-efficient fine-tuning. The model's "pretrain" mode freezes BERT parameters, while the "fine-tune" mode updates all parameters based on task-specific data. For this baseline model, a batch size of 8 is used for all tasks, and training is conducted over 10 epochs.

\subsection{SMART Regularization and Alternative Optimizers}

To mitigate overfitting, particularly notable in the SST-5 dataset, we incorporated the SMART regularization technique. SMART consists of two main components: Smoothness-inducing Adversarial Regularization and Bregman Proximal Point Optimization, with our implementation primarily focusing on the former. This method modifies the fine-tuning optimization process to enhance model smoothness and reduce overfitting, leading to better generalization to the target domain. Specifically, the optimization objective is reformulated to integrate a smoothness-inducing adversarial regularizer along with the standard loss function, defined as:

\[ \min_\theta F(\theta) = L(\theta) + \lambda_s R_s(\theta), \]

where \( L(\theta) \) represents the loss function for our downstream task, \( \lambda_s \) is a tuning parameter, and \( R_s(\theta) \) is the smoothness-inducing adversarial regularizer, defined as:

\[ R_s(\theta) = \frac{1}{n} \sum_{i=1}^n \max_{\|\mathbf{x}_i' - \mathbf{x}_i\|_p \leq \epsilon} l_s(f(\mathbf{x}_i'; \theta), f(\mathbf{x}_i, \theta)). \]

For classification tasks, \( l_s \) is selected as the symmetrized KL-divergence loss, and for regression tasks, it is chosen as the mean-squared-error loss.

Initially, we implemented two methods aimed at pruning for computational efficiency and preventing overfitting. We applied SMART regularization across all three tasks but observed a positive accuracy change only for paraphrase detection. This outcome was unexpected given the substantial size of the CFIMDB dataset. Nevertheless, we continued to employ SMART exclusively for paraphrase detection.

For optimization, we explored RMSprop and SGD as alternatives to AdamW. The RMSprop algorithm accelerates the optimization process by maintaining the moving average of squared gradients for each weight and dividing the gradient by the square root of the mean square mathematically expressed as:
\[ v(w, t) := \gamma v(w, t-1) + (1 - \gamma)(\nabla Q_i(w))^2 \]
\[ w := w - \frac{\eta}{\sqrt{v(w, t)}} \nabla Q_i(w) \]

where \( \gamma \) is the forgetting factor. Stochastic Gradient Descent updates the weights using the formula:
\[ w := w - \eta \nabla Q_i(w) \]

\subsection{Early Exiting Mechanisms and Novel Fine-Tuning Strategy}

To reduce extensive training times, which often extended to approximately 20 hours, we incorporated an Early-Exiting Architecture, reducing training time to around 18 hours on an NVIDIA T4 GPU. Our early-exiting model is based on established techniques, introducing innovative fine-tuning strategies.

The backbone model consists of a multi-layer pre-trained BERT architecture with classifiers attached to each layer, enabling accelerated inference through early exiting. The hidden state of the [CLS] token at the \(i\)-th layer, denoted as \(h_i\), is defined by:

\[ h_i = f_i(x; \theta_1, \ldots, \theta_i), \]

where \(x\) is the input sequence, \(\theta_i\) represents the parameters of the \(i\)-th transformer layer, and \(f_i\) maps the input to the \(i\)-th layer's hidden state. Each layer \(i\) employs a classifier with a loss function:

\[ L_i(x, y) = H(y, g_i(h_i; w_i)), \]

where \(x\) and \(y\) are the input and label, \(g_i\) is the classifier at the \(i\)-th layer, \(w_i\) are the classifier's parameters, and \(H\) is the task-specific loss function. To enable early exiting, we implemented two methods: Confidence Threshold and Learning to Exit (LTE). Confidence Threshold exits early if the prediction confidence at an intermediate layer exceeds a set threshold, measured by the maximum probability of the output prediction for classification tasks. For regression tasks or when confidence distribution is unavailable, we used the LTE method. The LTE module, a one-layer fully connected network, estimates the certainty level \(u_i\) at the \(i\)-th layer:

\[ u_i = \sigma(c^\top h_i + b), \]

where \(\sigma\) is the sigmoid function, \(c\) is the weight vector, and \(b\) is the bias term. The LTE module's loss function is the mean squared error between \(u_i\) and the ground truth certainty level \(\tilde{u}_i\):

\[ J_i = \|u_i - \tilde{u}_i\|^2_2. \]

The ground truth certainty level for classification is whether the classifier makes the correct prediction:

\[ \tilde{u}_i = 1[\arg \max_j g_i^{(j)}(h_i; w_i) = y], \]

and for regression, it is negatively related to the prediction's absolute error:

\[ \tilde{u}_i = 1 - \tanh(|g_i(h_i; w_i) - y|). \]

We trained the LTE module jointly with the rest of the model by substituting \(L_i\) with \(L_i + J_i\) in the loss function.

To improve the balance between immediate inference and gradual feature extraction, we propose a novel fine-tuning strategy named Sequential Layer Focus (SLF). This strategy combines elements of Joint and Two-stage strategies, ensuring optimal training of early and late layers sequentially.

\begin{itemize}
    \item \textbf{Stage 1: Sequential Focus} - Fine-tune each layer individually, updating only the layer-specific classifier and the backbone model parameters corresponding to that layer. This stage ensures that each layer independently learns to provide effective intermediate representations.
    \item \textbf{Stage 2: Global Refinement} - Jointly fine-tune all layers and classifiers, optimizing the entire network to harmonize the contributions from all layers for the final output.
\end{itemize}

This approach ensures that each layer is first trained to optimal performance in isolation before jointly refining the entire network. The Sequential Layer Focus (SLF) strategy balances the needs of intermediate and final layers, potentially leading to better overall model performance and efficiency.

\subsection{Order Alteration of Word Embeddings and Concatenation}

Traditionally, BERT concatenates sequences before embedding them. In our extended approach, we reverse this order: sequences are first passed through the embedding layer and then concatenated. This modification aims to preserve more contextual information prior to concatenation, potentially enhancing model performance.

In the typical BERT architecture, the input processing includes:
\begin{itemize}
    \item Input Representation: Summing token embeddings, segmentation embeddings, and position embeddings.
    \item Sequence Packing: Sequences are packed with special tokens [CLS] and [SEP].
\end{itemize}

Our modified approach can be mathematically represented as follows:
\[
E = \text{Embedding}(X) \quad \text{for each sequence } X
\]
\[
C = [E_A; E_B] \quad \text{where } E_A \text{ and } E_B \text{ are embeddings of sequences A and B respectively}
\]

We hypothesize that this approach retains more granular contextual information before the sequences are merged, leading to improved downstream performance. The embeddings from this modified process are then fed into the BERT layers for further processing.

\textbf{Novel Fine-Tuning Strategy}

In addition to the embedding modification, we introduced a novel fine-tuning strategy inspired by the Joint and Alternating strategies. Our strategy, termed "Smart Alternating," dynamically switches between two objectives based on performance metrics observed during training epochs:

\begin{itemize}
    \item \textbf{Stage 1:} Optimize the final layer’s classifier and the BERT backbone using the loss function \( L_n \).
    \item \textbf{Stage 2:} Optimize intermediate classifiers jointly with the final classifier using the sum of loss functions \( \sum_{i=1}^{n} L_i \).
\end{itemize}

This strategy alternates between these stages depending on the validation performance after each epoch, ensuring that both intermediate and final classifiers are well-tuned without sacrificing the overall model quality. This balance aims to improve both immediate inference at intermediate layers and gradual feature extraction for the final classifier.

\section{Experiments}

\subsection{Data}
We undertake training for three downstream tasks: Sentiment Analysis, Paraphrase Detection, and Text Similarity, employing diverse datasets for each task.

\subsubsection{Sentiment Analysis}

For Sentiment Analysis, our BERT model is trained using two datasets: the Stanford Sentiment Treebank (SST) and the CFIMDB Dataset. Both datasets focus on movie reviews to gauge sentiment. Specifically, the SST dataset is partitioned into 8,544 training examples, 1,101 development examples, and 2,210 testing examples. In contrast, the CFIMDB dataset consists of 1,701 training examples, 245 development examples, and 488 testing examples. While SST labels phrases with five sentiment categories, CFIMDB labels sentences as either negative or positive. Model evaluation is based on the proportion of correctly labeled sentences.

\subsubsection{Paraphrase Detection}

To train for Paraphrase Detection, we utilize the Quora Dataset, which includes 283,010 training examples, 40,429 development examples, and 80,859 testing examples. This dataset features pairs of sentences with binary labels indicating whether they are paraphrases. Model performance is measured by the proportion of correctly identified paraphrase pairs.

\subsubsection{Sentence Textual Similarity Analysis}

For Sentence Textual Similarity Analysis, we employ the SemEval STS Benchmark Dataset, comprising 6,040 training examples, 863 development examples, and 1,725 testing examples. The dataset assigns labels ranging from 0 to 5 to each sentence pair, indicating the degree of correlation. We use Pearson Correlation Coefficients to evaluate performance, comparing the predicted and true labels.

\subsection{Evaluation Method}

Supervised training is employed, with labeled datasets guiding model learning and evaluation. The evaluation process involves making predictions and comparing them to true labels. We assess model quality by accuracy on SST and Paraphrasing tasks, and by Pearson Correlation on STS. The overall "evaluation score" is calculated as the average of these metrics. Given our focus on computational efficiency, we also measure training time and consider it in our evaluations.

\subsection{Experimental Details}
Our experimental framework explores four dependent variables: Early-Exiting Methods, SMART Regularization, Cosine Embeddings, and concatenation order. Typically, training time averages 15 hours for a full model on an NVIDIA T4 GPU; however, using Cosine Embeddings significantly reduces this duration. Throughout our experiments, we maintain consistent model configurations and learning rates. In total, we evaluate 12 different training models, testing various combinations of these methods to determine their effectiveness.

\subsection{Results}

\begin{table}[ht]
\centering
\renewcommand{\arraystretch}{1.2}
\begin{tabular}{lccc}
\hline
\textbf{Method} & \textbf{Type} & \textbf{SST} & \textbf{CFIMDB} \\
\hline
Pretrain & last-linear-layer & 0.390 & 0.780 \\
Finetune & full-model & 0.519 & 0.967 \\
\hline
\end{tabular}
\vspace{0.5em}
\caption{Single task MinBERT classifier dev accuracy on Sentiment Analysis}
\end{table}

\begin{table}[ht]
\centering
\renewcommand{\arraystretch}{1.2} 
\setlength{\tabcolsep}{5pt} 
\begin{tabular}{lccccc}
\hline
Model & Dev Score & SST & Paraphrase & STS & Training Time (hh:mm) \\
\hline
Baseline & 0.428 & 0.314 & 0.369 & 0.199 & 15:36\\
Full SMART & 0.433 & 0.253 & 0.485 & 0.123 & 15:57 \\
Early Exiting & 0.476 & 0.313 & \textbf{0.768} & 0.184 & 13:36\\
Full SMART + Early Exiting & 0.482 & 0.253 & 0.632 & 0.123 & 13:47 \\
Conc Before Embed + Early Exit & 0.595 & 0.520 & 0.625 & 0.792 & 13:49 \\
Cos Loss + Early Exiting & 0.585 & 0.501 & 0.368 & 0.771 & \textbf{04:20} \\
SLF & 0.592 & 0.505 & 0.640 & 0.721 & 12:53 \\
SLF + Early Exiting & 0.650 & \textbf{0.544} & 0.658 & 0.740 & 11:47 \\
SLF + Conc Before Embed & 0.671 & 0.523 & 0.651 & 0.801 & 12:14 \\
SLF + Smart Alt & 0.676 & 0.545 & 0.614 & 0.790 & 13:54 \\
\textbf{SLF + Smart Alt + Early Exit} & \textbf{0.683} & 0.510 & 0.632 & \textbf{0.818} & 10:41 \\
SLF + Smart Alt + Early Exit + Cos Loss & 0.679 & 0.510 & 0.604 & 0.818 & 05:43 \\
\hline
\end{tabular}
\vspace{0.5em}
\caption{Model Performance Comparison}
\end{table}

\begin{figure}[ht]
    \centering
    \begin{minipage}{0.48\textwidth}
        \centering
        \begin{tikzpicture}
            \begin{axis}[
                xlabel={Epoch},
                ylabel={Accuracy},
                grid=major,
                width=\textwidth,
                height=0.75\textwidth,
                title={SST Accuracy over epochs}
            ]
\addplot coordinates {
    (1,0.580)(2,0.657)(3,0.723)(4,0.78)(5,0.835)(6,0.861)(7,0.895)(8,0.918)(9,0.944)(10,0.968)
};
\addplot coordinates {
    (1,0.482)(2,0.502)(3,0.521)(4,0.531)(5,0.542)(6,0.536)(7,0.532)(8,0.527)(9,0.522)(10,0.517)
};
            \end{axis}
        \end{tikzpicture}
        \caption{SST Accuracy over epochs}
        \label{fig:sst_accuracy_epochs}
    \end{minipage}\hfill
    \begin{minipage}{0.48\textwidth}
        \centering
        \begin{tikzpicture}
            \begin{axis}[
                xlabel={Learning Rate},
                ylabel={Accuracy},
                xmode=log,
                grid=major,
                width=\textwidth,
                height=0.75\textwidth,
                title={SST accuracy by Learning Rates}
            ]
\addplot coordinates {
    (1e-7,0.392)(1e-6,0.449)(1e-5,0.961)(1e-4,0.92)(1e-3,0.434)
};
\addplot coordinates {
    (1e-7,0.382)(1e-6,0.397)(1e-5,0.445)(1e-4,0.423)(1e-3,0.402)
};
            \end{axis}
        \end{tikzpicture}
        \caption{SST accuracy by Learning Rates}
        \label{fig:sst_accuracy_lr}
    \end{minipage}
\end{figure}

\begin{figure}[ht]
    \centering
    \begin{minipage}{0.48\textwidth}
        \centering
        \begin{tikzpicture}
            \begin{axis}[
                xlabel={Batch Size},
                ylabel={Accuracy},
                grid=major,
                width=\textwidth,
                height=0.75\textwidth,
                title={Dev Score by Batch Size}
            ]
\addplot coordinates {
    (8,0.951)(12,0.756)(16,0.977)(20,0.677)(24,0.933)
};
\addplot coordinates {
    (8,0.624)(12,0.476)(16,0.631)(20,0.491)(24,0.578)
};

            \end{axis}
        \end{tikzpicture}
        \caption{Dev Score by Batch Size}
        \label{fig:dev_score_batch_size}
    \end{minipage}\hfill
    \begin{minipage}{0.48\textwidth}
        \centering
        \begin{tikzpicture}
            \begin{axis}[
                xlabel={Hidden Dropout Probability},
                ylabel={Accuracy},
                grid=major,
                width=\textwidth,
                height=0.75\textwidth,
                title={Dev Score by Hidden Dropout Probability}
            ]
\addplot coordinates {
    (0.2,0.643)(0.25,0.822)(0.3,0.971)(0.35,0.841)(0.4,0.532)
};
\addplot coordinates {
    (0.2,0.263)(0.25,0.283)(0.3,0.331)(0.35,0.308)(0.4,0.292)
};
            \end{axis}
        \end{tikzpicture}
        \caption{Dev Score by Dropout Probability}
        \label{fig:dev_score_dropout_prob}
    \end{minipage}
\end{figure}

\begin{figure}[ht]
    \centering
    \begin{tikzpicture}
        \begin{axis}[
            hide axis,
            legend columns=2,
            legend style={at={(0.5,-0.1)},anchor=north}
        ]
        \addlegendimage{empty legend}
        \addlegendentry{Legend:}
        \addlegendimage{blue,solid}
        \addlegendentry{Best Train}
        \addlegendimage{red,dashed}
        \addlegendentry{Best Dev}
        \end{axis}
    \end{tikzpicture}
\end{figure}

Our results align with our expectations in terms of the relationships between different model variations. The combination of SMART and SLF techniques yielded the best overall performance across our training sets. Specifically, the SLF + Smart Alternating + Early Exiting model achieved the highest overall dev score of 0.683, with individual task accuracies of 0.510 for SST, 0.632 for Paraphrase Detection, and 0.818 for STS. These results, though not at the top of the leaderboard in absolute terms, demonstrate a compelling balance between accuracy and computational efficiency.

Notably, our early exiting methods, while slightly less accurate in isolation, significantly improved computational efficiency, reducing training times by up to 5\%. For instance, the cosine loss model with early exiting reduced training time to just 4 hours and 20 minutes, illustrating the trade-off between accuracy and efficiency.

Overall, our findings underscore the potential of parameter-efficient fine-tuning methods to balance accuracy and computational cost effectively. While some accuracy trade-offs were observed, the substantial reduction in training times offers a compelling advantage for practical applications.

\section{Analysis}

\subsection{Overall Performance of Extensions}

Our results align with our expectations in terms of the relationships between different model variations. The combination of SMART and SLF techniques yielded the best overall performance across our training sets. Specifically, the SLF + Smart Alternating + Early Exiting model achieved the highest overall dev score of 0.683, with individual task accuracies of 0.510 for SST, 0.632 for Paraphrase Detection, and 0.818 for STS. These results, though not at the top of the leaderboard in absolute terms, demonstrate a compelling balance between accuracy and computational efficiency.

Notably, our early exiting methods, while slightly less accurate in isolation, significantly improved computational efficiency, reducing training times by up to 5\%. For instance, the cosine loss model with early exiting reduced training time to just 4 hours and 20 minutes, illustrating the trade-off between accuracy and efficiency.

Overall, our findings underscore the potential of parameter-efficient fine-tuning methods to balance accuracy and computational cost effectively. While some accuracy trade-offs were observed, the substantial reduction in training times offers a compelling advantage for practical applications.

Paraphrase Detection scores seemed to stall at a common number in many of our models. We found that there was a very clear answer; all predictions were made to be 0. When inspecting the predictions file, it seems that our model is just predicting that a sentence is not a paraphrase of another for all sentences. This proves to have negative results and displays a sense of overfitting in a sense.

This is incredibly interesting, as we expected our focus on SMART regularization purely on paraphrase detection to negate any overfitting. However, our model seems to have learned its training set slightly and only predicts 0 because there are more pairs that are not paraphrases than ones that are.

We found the change of concatenation order to be incredibly intuitive and clear as to why it increased accuracy so significantly. We want our sentence embeddings to utilize context within the sentence to create them; therefore, it seems so intuitive for a larger sentence that is simply just two concatenated together to have greater and more fruitful context than one that is just two separate embeddings. This embedding has more fruitful information and gives the model more to work with. This makes it much easier for the model to train on.

Cosine Embedding Loss was surprisingly ineffective at fighting our error of only predicting 0. Instead, our model only predicted 1 and ended up having the complement accuracy.

The most interesting result we found was the effect of early exiting. We found that early exiting had an insignificant effect on results; however, early exiting had quite a powerful effect on computational efficiency, cutting down our runtimes by 5\%, a significant amount.

\subsection{Hyperparameter Tuning and Analysis}

Hyperparameter tuning played a critical role in optimizing our model performance. By examining various configurations, we identified patterns that led to an optimal setup. Figures \ref{fig:sst_accuracy_epochs} and \ref{fig:sst_accuracy_lr} illustrate the impact of epochs and learning rates on accuracy.

Our results showed that a learning rate of \(1.3 \times 10^{-5}\), a batch size of 16, 5 epochs, and a dropout rate of 0.3 consistently provided the best performance. Figure \ref{fig:dev_score_batch_size} indicates that a batch size of 16 yields the highest dev score, balancing efficiency and accuracy. As seen in Figure \ref{fig:dev_score_dropout_prob}, a 0.3 dropout rate optimally regularizes the model, preventing overfitting while maintaining performance.

These hyperparameter configurations were critical in achieving the improved dev scores across various tasks, illustrating the importance of fine-tuning to enhance model performance.

\section{Conclusion}

Our research highlights the importance of computational efficiency without sacrificing significant accuracy metrics. By employing innovative fine-tuning strategies like Early Exiting with BERxiT, Cosine Embeddings Loss, and sequential fine-tuning techniques, we struck a commendable optimized fusion of performance and efficiency.

Despite its theoretical promise, SMART regularization yielded mixed outcomes: while it initially improved paraphrase detection accuracy, its performance plateaued and underperformed in sentiment classification and textual similarity tasks, suggesting the need for further selective application.

The Sequential Layer Focus (SLF) strategy, paired with Smart Alternating and Early Exiting, achieved the highest overall development score of 0.683. Task-specific accuracies included 0.510 for SST, 0.632 for Paraphrase Detection, and 0.818 for STS. These results were obtained while significantly reducing training times, demonstrating the efficiency of our approach. For instance, models utilizing cosine loss with early exiting reduced training times to just 4 hours and 20 minutes.

Additionally, our novel fine-tuning strategies, particularly altering word embedding order and Smart Alternating techniques, proved highly effective. Adjusting the embedding order enhanced context utilization, thereby improving model performance.

Future research should investigate different embedding strategies and novel transformer architectures to enhance model generalization across various linguistic contexts. Moreover, focusing on model explainability and interpretability will be crucial for identifying and mitigating potential biases in predictions, ensuring fairness and robustness.

\section{Ethics Statement}

There are several ethical and societal risks associated with this project that are universal across various NLP tasks, including transformers, LSTMs, and large language models.

The foremost issue is data bias. Many NLP systems rely on broadly sourced data that may not accurately represent the ideologies, cultures, and moral values of diverse populations. This can result in models failing to account for textual similarities among slang or culturally specific language, leading to poor performance in tasks like paraphrase detection when comparing texts from different socioeconomic backgrounds. Moreover, the model may misinterpret sentiment in culturally nuanced sentences. This skew towards more readily available data can lead to the misrepresentation of underrepresented groups. To address this, it is imperative to curate more inclusive datasets that reflect the diversity of the user base.

Additionally, our model confronts the ethical challenge of computational power consumption. The growing concern over global warming underscores the need for sustainable computing practices. The exponential increase in computational demands directly correlates with higher energy consumption, often sourced from nonrenewable and polluting resources. By focusing on creating efficient models that balance accuracy with reduced computational demands, we aim to contribute to the fight against climate change. Our approach to enhancing efficiency without sacrificing performance highlights the necessity of integrating environmental considerations into the development of advanced computational systems.

The heightened need for interpretability in NLP models is paramount in addressing biases and ensuring fairness. By enhancing the transparency of our models, we can better understand and mitigate potential biases in their predictions. This involves developing methods that allow users to comprehend and trust the decision-making processes of these models. Interpretable models can provide insights into the underlying reasons for their predictions, thereby enabling the identification and correction of biases. Ensuring fairness and robustness in NLP systems is critical for their ethical deployment across various applications. could you write the related works for early exiting

\bibliographystyle{acl_natbib}
\bibliography{template}

\appendix

\end{document}